\documentclass[runningheads]{llncs}
\usepackage{graphicx}
\usepackage{tikz}
\usepackage{comment}
\usepackage{amsmath,amssymb} % define this before the line numbering.
\usepackage{color, colortbl}
\usepackage{xcolor}

\usepackage{booktabs, array}
\usepackage{subcaption}
\usepackage{multirow}
\usepackage{multicol}
\usepackage{makecell}
\usepackage[toc, page]{appendix}
\usepackage{float}

\usepackage[accsupp]{axessibility}  % Improves PDF readability for those with disabilities.

\usepackage[pagebackref,breaklinks,colorlinks]{hyperref}

\begin{document}
% \renewcommand\thelinenumber{\color[rgb]{0.2,0.5,0.8}\normalfont\sffamily\scriptsize\arabic{linenumber}\color[rgb]{0,0,0}}
% \renewcommand\makeLineNumber {\hss\thelinenumber\ \hspace{6mm} \rlap{\hskip\textwidth\ \hspace{6.5mm}\thelinenumber}}
% \linenumbers
\pagestyle{headings}
\mainmatter

\title{A Simple Baseline for \\
Low-Budget Active Learning} % Replace with your title

% INITIAL SUBMISSION 
\begin{comment}
\titlerunning{ECCV-22 submission ID \ECCVSubNumber} 
\authorrunning{ECCV-22 submission ID \ECCVSubNumber} 
\author{Anonymous ECCV submission}
\institute{Paper ID \ECCVSubNumber}
\end{comment}
%******************

% CAMERA READY SUBMISSION
% \begin{comment}
\titlerunning{~}
% If the paper title is too long for the running head, you can set
% an abbreviated paper title here
%
% \author{\fontsize{11}{11} \selectfont Kossar Pourahmadi$^{1} \quad$ 
%  Parsa Nooralinejad$^{1} \quad$ Hamed Pirsiavash$^{1,2}\quad$   \\
%  \\
 
% \fontsize{11}{11} \selectfont $^{1}$ University of Maryland, Baltimore County $\quad$  $^{2}$University of California, Davis\\
% {\tt\fontsize{9}{9} \selectfont \{kossarp1, parsan1\}@umbc.edu \quad hpirsiav@ucdavis.edu}

% }
\author{Kossar Pourahmadi$^{1}$ \quad
Parsa Nooralinejad$^{1}$ \quad
Hamed Pirsiavash$^{2}$}
\authorrunning{~}
% First names are abbreviated in the running head.
% If there are more than two authors, 'et al.' is used.
%
\institute{$^{1}$University of Maryland, Baltimore County \quad $^{2}$University of California, Davis \\
\email{\{kossarp1, parsan1\}@umbc.edu} \quad \email{hpirsiav@ucdavis.edu}}
% \end{comment}
%******************
\maketitle
\begin{abstract}
Active learning focuses on choosing a subset of unlabeled data to be labeled. However, most such methods assume that a large subset of the data can be annotated. We are interested in low-budget active learning where only a small subset ({\em e.g.}, $0.2\%$ of ImageNet) can be annotated. Instead of proposing a new query strategy to iteratively sample batches of unlabeled data given an initial pool, we learn rich features by an off-the-shelf self-supervised learning method only once, and then study the effectiveness of different sampling strategies given a low labeling budget on a variety of datasets including ImageNet. We show that although the state-of-the-art active learning methods work well given a large labeling budget, a simple $K$-means clustering algorithm can outperform them on low budgets. We believe this method can be used as a simple baseline for low-budget active learning on image classification. Code is available at:~\href{https://github.com/UCDvision/low-budget-al}{https://github.com/UCDvision/low-budget-al}

\end{abstract}

\section{Introduction}

\begin{figure*}[t]
\captionsetup[subfigure]{justification=centering}
\centering
\includegraphics[width=.9\linewidth]{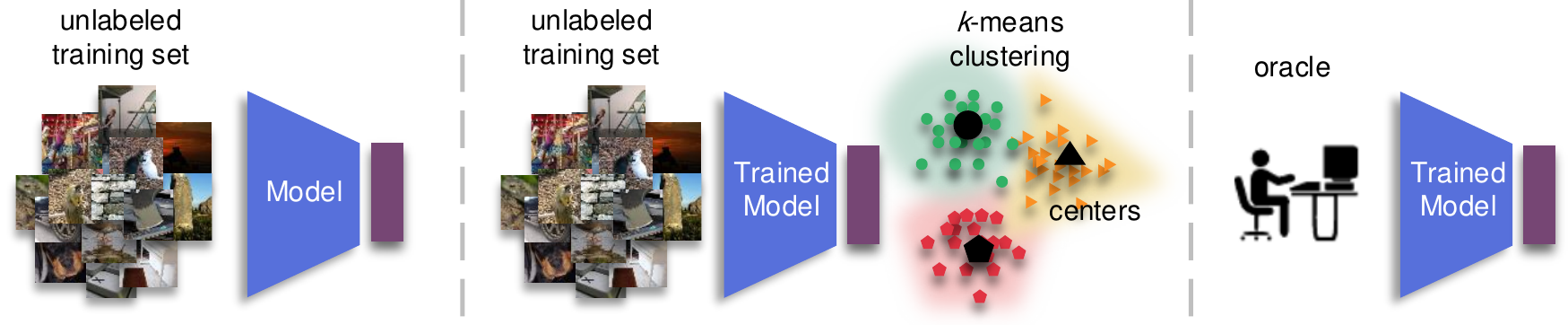}
    \begin{minipage}[t]{.3\linewidth}
    \centering
    \subcaption{Self-supervised training}\label{fig:overview:1}
    \end{minipage}
    \begin{minipage}[t]{.3\linewidth}
    \centering
    \subcaption{Feature extraction and clustering}\label{fig:overview:2}
    \end{minipage}
    \begin{minipage}[t]{.3\linewidth}
    \centering
    \subcaption{Annotation and fine-tuning}\label{fig:overview:3}
    \end{minipage}

   \caption{{\bf Our simple baseline.} The goal is to train an accurate image classification model with a very small set of labeled images. {\bf a)} A self-supervised learning model learns from the unlabeled data and provides the feature embeddings. {\bf b)} $K$-means algorithm clusters the unlabeled data features and chooses data points nearest to the center of each cluster. {\bf c)} These selected points are then annotated by an oracle. Finally, a linear classifier on the top of features learns from the annotated data and performs the image classification task.}
\label{fig:overview}
\vspace{-.15in}
\end{figure*}

We are interested in active learning with very low budget. Given a large set of unlabeled images and an oracle that can label a small set of images, we want to train an accurate image classification model by choosing a small set of images for the oracle to annotate. Active learning \cite{al1} has been studied for a long time. However, most active learning methods start with a large initial seed pool (usually randomly chosen images that are annotated) and also choose a large set of images to be annotated actively. For instance, \cite{VAAL,MAL,ensemble} use more than $3\%$ of ImageNet \cite{imagenet} as the initial pool, which contains more than $40,000$ images. 
% For instance, VAAL \cite{VAAL} uses $10\%$ of ImageNet \cite{imagenet} as the initial pool, which contains more than $120,000$ images. 
In some applications, {\em e.g.}, medical image analysis, the labeling budget is much smaller, so the current active learning algorithms may not be a viable solution \cite{kuo2018cost,yang2017suggestive}. 

One may argue that our setting is similar to few-shot and semi-supervised learning where we assume a few examples of each class are labeled. However, we argue that those settings are not practical in many applications since sampling a subset of images for labeling, that are  ``uniformly'' distributed across categories, itself needs a larger subset of data to be annotated first. In some real applications, even finding one example of an object to annotate may be challenging. For instance, in a self-driving car application, one may want to build a detector for motorbikes with annotating a few motorbike examples only, but finding those few examples in the large dataset of many hours of driving video footage is a challenging task by itself. 

As an example, Table \ref{tab:cat_cov_imagenet} shows that some categories may not be even represented in a pool of $3,000$ randomly sampled ImageNet images. This makes standard few-shot or semi-supervised learning methods impractical. Hence, we believe active learning, when we want to have only a few annotations per category in average, is an important practical problem. 

\begin{figure*}[t]
\centering
\includegraphics[width=1\textwidth]{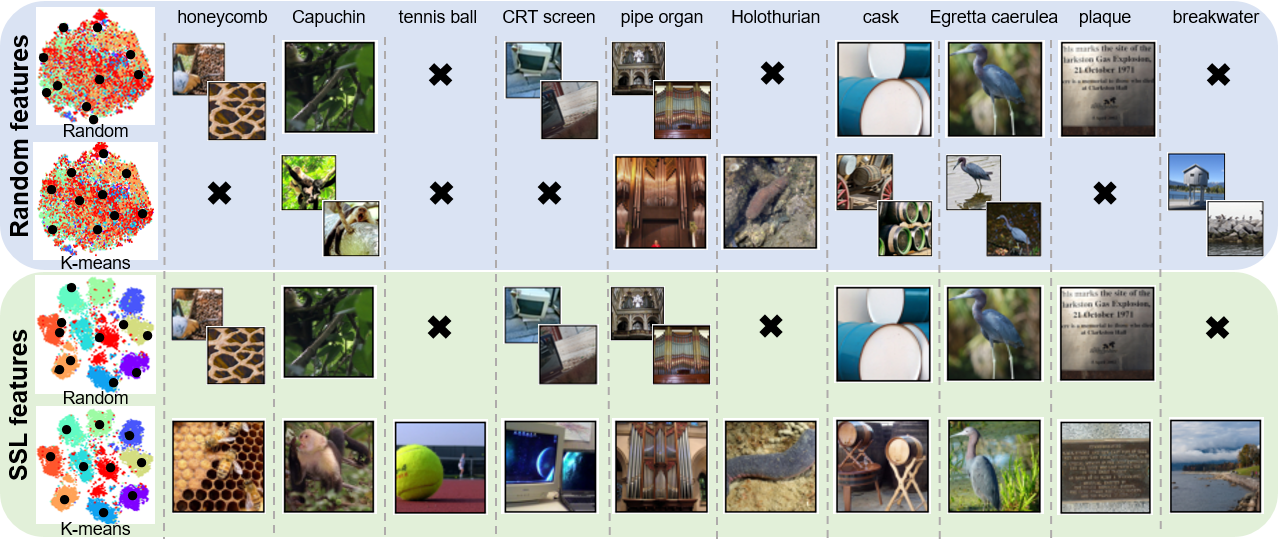}
   \caption{{\bf Visualization of category coverage of $\textbf{10}$ samples on $\textbf{10}$ randomly selected ImageNet categories.}
   We show t-SNE plots of two different feature initializations (random and SSL) and selection methods (Random and $K$-means) along with $10$ selected images (black dots) for annotation. %This is done on samples of only $10$ randomly selected ImageNet categories.
   %(honeycomb, Capuchin, tennis ball, CRT screen, pipe organ, Holothurian, cask, Egretta caerulea, plaque, and breakwater). 
   We have not done any cherry-picking or manual inspection for this visualization. Note that random selection uses the same seed, so Rows $1$ and $3$ have the same selected images. SSL initialization along with $K$-means results in $100\%$ coverage while some categories are not represented in other settings. Table \ref{tab:cat_cov_imagenet} shows that on full ImageNet, our method covers $99.9\%$ of all categories with $7$ samples on average per class.}
\label{fig:feat_space}
\end{figure*}

We believe given the recent progress in self-supervised learning (SSL) \cite{compress,simclr,MoCo-v2}, active learning with very small budget ({\em e.g.}, annotating only $3,000$ images of ImageNet) should have become more tractable. Hence, we design a very simple baseline and show that it outperforms state-of-the-art active learning methods 
% by a large margin 
on small budgets. As shown in Figure \ref{fig:overview}, our baseline starts with training an off-the-shelf SSL method on the unlabeled data, running $K$-means clustering on the obtained features of unlabeled data, choosing examples closest to the center of each cluster, annotating those samples using the oracle, and finally training a linear classifier on the top of SSL features to perform the image classification task. Figure \ref{fig:feat_space} shows how using $K$-means selection on SSL pre-trained features nicely covers categories of the unlabeled data.

Moreover, most active learning methods \cite{coreset,gao2020consistency,VAAL} annotate multiple batches of data sequentially which constrains the annotation work-flow since the oracle should wait for the model to be trained and select the next batch. We show that a variation of our method works well using a single annotation batch rather than multiple ones, reducing the cost of annotation work-flow.

%should wait for the next model to be trained. We demonstrate that a variation of our method, that performs the best annotates only a single selected batch of data rather than iteratively annotating multiple batches, which is common in active learning \cite{coreset,gao2020consistency,VAAL}. Single-batch sampling reduces the annotation costs by simplifying the annotation workflow and not having the oracle wait for the machine to select the next batch.

%%%%%%%%

\section{Related Work}

There is a broad array of learning approaches to give deep learning-based methods an ability of learning new concepts without requiring large annotated datasets. These approaches include few-shot learning, active learning, semi-supervised learning, and self-supervised learning. 
\vspace{.05in}

\noindent{\bf Few-shot learning.} Few-shot learning aims to recognize a set of classes that there are very few uniformly distributed training examples available for each of them ({\em e.g.} 1 or 5 examples per class) \cite{ll,cc,proto}. Although the amount of few-shot learning training data is much less than of large-scale deep learning methods, preparing a training set with as few as $1\%$ uniform annotation still needs more than $1\%$ of the unlabeled dataset to be annotated.

\vspace{.05in}
\noindent{\bf Active learning.} Active learning has been widely studied to answer {\em how} to choose a fixed number of samples to gain the highest accuracy. Our work lies in pool-based approaches which can be categorized as uncertainty-based and distribution-based methods. Uncertainty-based methods try to find the most uncertain samples \cite{entropy,tong2001support,brinker2003incorporating,aggarwal2022optimizing}. In Bayesian uncertainty-based approaches, Gaussian processes estimate uncertainty \cite{gaussian,mc,mcdropout}. However, it is shown in \cite{coreset} that these methods are not suitable for deep neural networks and large-scale datasets.

Distribution-based methods try to maximize the diversity of selected data over the entire dataset. \cite{pre-clustering} clusters the dataset at first and then uses a hierarchical approach to avoid selecting repeated samples from the same cluster. Core-set \cite{coreset} increases diversity in the selected batch of data by minimizing the Euclidean distance between instances of already labeled data and instances in the unlabeled pool. Despite working well on datasets with small number of classes, the performance of Core-set degrades when {\em p}-norms suffer from the curse of dimensionality in high-dimensions \cite{franccois2008high}.

Some methods take advantage of both uncertainty and diversity \cite{adaptive,batchbald,badge}. VAAL \cite{VAAL} and DFAL \cite{dfal} use adversarial learning to learn the representation of data points. It is shown in MAL \cite{MAL} that although VAAL does not need the annotations to sample data, it may result in selecting multiple instances of the same class while there are already plenty of them in the labeled pool.
%  MAL tries to solve this problem by ranking the unlabeled data with respect to the similarity to already labeled samples. 

Many active learning methods require a large initial labeled pool and sampling budget \cite{coreset,VAAL,MAL,pmlr-v108-shui20a}. This condition is difficult to meet in medical image analysis or other domains that the size of unlabeled training data is very large and it costs experts a lot of effort to label a large subset of them \cite{hoi2006batch}. An approach to mitigate this problem is presented in \cite{mahmood2021low}. However, it requires computing large distance matrices to solve Wasserstein distances for large datasets ({\em e.g.}, with $1$M images) that presents a scalability bottleneck.
% solving Wasserstein distances for large datasets ({\em e.g.}, with $1$M images) requires computing large distance matrices and presents a bottleneck. 

In this paper, we train an off-the-shelf SSL method on the unlabeled data to provide rich feature vectors only once and select a small proportion of them in a non-iterative approach to achieve strong performance on a variety of datasets, as well as large-scale datasets, without requiring an initial labeled pool. 

% \vspace{.05in}
\noindent{\bf Semi-supervised learning.} Semi-supervised learning for image classiﬁcation aims in making best use of limited labeled data while leveraging the unlabeled dataset \cite{simclr-v2,NEURIPS2020_44feb009,kuo2020featmatch}.
% leveraging unlabeled data and improving model performance by
Three most explored directions are consistency regularization \cite{NIPS2016_30ef30b6,LaineA17}, entropy minimization \cite{NIPS2004_96f2b50b}, and pseudo-labeling \cite{mixmatch,NEURIPS2020_06964dce}. Despite achieving highly competitive performance to supervised methods, semi-supervised learning, similar to few-shot learning, assumes there are a few equal number of examples per category and needs more than the amount of training set to be annotated from the unlabeled dataset to ensure a uniform distribution. Therefore, we cannot compare active and semi-supervised methods directly. However, in Section \ref{semi-sup}, we will introduce two labeled set generation settings to show that active strategies can benefit the performance of semi-supervised models.

\noindent{\bf Self-supervised learning.} SSL provides a strategy to train a neural network on the unlabeled data and creates rich features that can be fine-tuned for different downstream tasks using limited labeled data. SSL methods coarsely learn to solve a pretext task \cite{rotpred,counting,jigsaw} or contrast between similar pairs of an image and its negative pairs \cite{swav,simclr,MoCo}.
%  A contrastive learning method enforces representations to maximize agreement between different transformations of the same image and minimize it between an image and other images \cite{hadsell2006dimensionality}.
CompRess \cite{compress} is one of the recent state-of-the-art SSL models that compresses a deep teacher network to a smaller student network such that for any query, the student ranks anchors similar to the teacher.

Prior active learning methods that take advantage of unsupervised learning tackle with time-consuming human-in-the-loop workflow of waiting for the new model to train on previous batches to annotate a new selected batch \cite{simeoni2019rethinking,gao2020consistency,zhang2020state,MAL}, or use large sampling budgets \cite{chandra2020initial,mottaghi2019adversarial,emam2021active}. In contrast, we select a single batch of a few examples using the initial SSL pre-trained model.

%%%%%%%%
\section{Low-Budget Active Learning}
\subsection{Standard Active Learning}
% In this section, we set up notations and define the scenario of active learning for the rest of the paper. 

Assuming a fixed sampling budget $N$ and a large pool of unlabeled data $D_u$, standard active learning algorithms train a model $\mathcal{M}$ using an initial labeled pool $D_l^0$. Then, at iteration $t$, they sample a subset $D_l^t$ of images from $D_u$ to be labeled manually. This subset expands the labeled data to $D_l = \cup_t D_l^{t}$. The number of iterations and size of each subset are chosen so that $|D_l|$ matches the available budget $N$. The final model $\mathcal{M}$ is trained on all labeled data.
\vspace{-.1in}
\subsection{Sampling with $K$-means Clustering Method}
In this paper, we use two forms of $K$-means sampling including: $i$) {\em single-batch $K$-means} and $ii$) {\em multi-batch $K$-means}.

\vspace{.05in}
\noindent{\bf Single-batch K-means.} Since we are interested in the low-budget setting, we eliminate the need for initial seed of labeled samples and in contrast to standard iterative active learning methods, we perform only one iteration of sampling. We simply apply $K$-means algorithm on feature outputs of an SSL pre-trained model only once and choose samples closest to the cluster centers. The single-batch method simplifies the annotation workflow as the annotators do not need to wait for the new model to train before annotating a new batch.
However, one may need to know the whole number of samples that need annotation. We call single-batch $K$-means simply as {\em $K$-means} in the rest of the paper.

\vspace{.05in}
\noindent{\bf Multi-batch K-means.} In contrast to single-batch version, multi-batch $K$-means is dependent on the previous sampling rounds. This method uses the difference of two consecutive budget sizes as the number of clusters and picks those nearest examples to centers that have not been labeled previously by the oracle. We call this variant of $K$-means sampling as {\em multi $K$-means}. Although we no longer waste previously labeled data and accumulate them to the new set of labeled examples, this process is iterative and we should wait for all previous rounds to finish. As a result, both scenarios (single-batch and multi-batch) have their own advantages and disadvantages and one can choose the method that fits the best to the application setting.

%%%%%%%%%%%
\vspace{-.1in}
\section{Experiments and Results}
\label{exp}

In this section, we evaluate different active learning baselines on image classiﬁcation task.

\vspace{.05in}
\noindent{\bf Datasets.} We use CIFAR-10/100 \cite{krizhevsky2009learning}, ILSVRC-2012 ImageNet \cite{imagenet}, and Im-ageNet-LT \cite{imagenet_lt} to compare different sampling strategies. CIFAR-10/100 with $10$/$100$ categories have $60,000$ images of size $32\times32$, where $50,000$ are training and $10,000$ are test images. ImageNet contains more than $1.2$M images that are almost uniformly distributed over $1,000$ categories. ImageNet-LT is truncated from ImageNet and has the same $1,000$ categories, but the number of images per class ranges from 1280 to 5. We resize ImageNet/LT images into $224\times224$ pixels. The validation set for ImageNet/LT experiments is the same and contains $50,000$ samples. All datasets are augmented by horizontally flipping the images.

% We use CIFAR-10/100 \cite{krizhevsky2009learning} with $10$/$100$ classes and $60$k images, ImageNet \cite{imagenet} with $1$k classes and $1.3$M images, and ImageNet-LT \cite{imagenet_lt} with a non-uniformly distributed subset of ImageNet images over $1$k classes.

\vspace{.05in}
\noindent{\bf Baselines.} We compare {$K$}-means and multi {$K$}-means with the following baselines: {$i$}) {\bf Random} in which samples are selected randomly (uniformly) from the entire dataset; {$ii$}) {\bf Max-Entropy} \cite{entropy} which samples the points with the highest probability distribution entropy; {$iii$}) {\bf Core-set} \cite{coreset}; {$iv$}) {\bf VAAL} \cite{VAAL}; and {$v$}) {\bf Uniform} that selects equal number of random samples per class. Note that few-shot and semi-supervised learning methods use Uniform strategy to create training sets that may require more than the size of sets to be annotated.

% \vspace{.05in}
\noindent{\bf Implementation details.} In all experiments, unless specified, we use feature outputs of ResNet-18 that is pre-trained on unlabeled ImageNet using CompRess SSL method \cite{compress} for $130$ epochs, which uses MoCo-v2 \cite{MoCo-v2} as its teacher network.
% Core-set, multi {$k$}-means, and {$k$}-means use feature outputs of ResNet-18 that is pre-trained on unlabeled ImageNet using CompRess SSL method \cite{compress} for $130$ epochs, which uses MoCo-v2 \cite{MoCo-v2} as its teacher network. 
Note that this pre-trained feature extractor is used even for CIFAR experiments which means that technically, CIFAR experiments use the unlabeled data beyond CIFAR datasets. Max-Entropy sampling method \cite{entropy} freezes the pre-trained backbone and trains an extra linear layer as the classifier on the top of that for $100$ epochs. In all Max-Entropy experiments, we use Adam optimizer and lr=$0.001$ which is multiplied by $0.1$ at epochs $50$ and $75$. The budget for iterative sampling methods is the difference between two consecutive budget sizes. For Random, Uniform, and {$K$}-means sampling, each budget size is equivalent to the amount of unlabeled data selection.

\noindent{\bf Evaluation metrics.} 
Unless specified, all experiments are averaged over $3$ runs with $3$ constant random seeds. We follow four evaluation metric protocols:
% We follow four evaluation metric protocols to compare the performance of different active learning methods. 

\noindent{$i$}) {\bf Linear classification.} We train a linear classifier on the top of the frozen backbone features (without back-propagation in the backbone weights) on the pool of labeled data for $100$ epochs and report its top-1 accuracy on the test set. We apply the mean and standard deviation normalization at each dimension of backbone outputs to reduce the computational overhead of tuning the hyper-parameters per experiment. We use Adam optimizer and lr=$0.01$ that is multiplied by $0.1$ at epochs $50$ and $75$. The batch size is $128$ on ImageNet/LT. For CIFAR-10/100 experiments, initial pools contain only 10/100 examples, so we set batch size=$4$. 

\noindent{$ii$}) {\bf Nearest neighbor classification.} This uses cosine similarity as a distance metric to search for the most semantically similar neighbors of test set data from the pool of labeled images. When the pool of labeled data is small, this metric is faster than linear evaluation since nearest neighbor classification needs no hyper-parameter tuning. We use FAISS GPU library \cite{faiss} for implementation.

\noindent{$iii$}) {\bf Evaluation on fine-grained tasks.} We evaluate on Flowers-102 \cite{nilsback2008automated}, DTD-47 \cite{cimpoi2014describing}, and Aircraft \cite{maji2013fine} as examples of fine-grained datasets. For feature embeddings, we use the same frozen backbone that is pre-trained on unlabeled ImageNet. 
In each dataset, we first choose a small subset of training data to be annotated, then we use the subset to train a linear classifier on top of the frozen backbone. This is similar to the transfer learning procedure in \cite{simclr,byol}. Full details on training sets are in the appendix Table \ref{tab:appendix_transfer_dset_details}.

% We choose a small subset of the fine-grained datasets to be manually annotated, and then use them to train a linear classifier. This is similar to the procedure in \cite{simclr,byol}.

%We follow the procedure in \cite{simclr,byol} to perform transfer learning by choosing a small subset of the fine-grained datasets to be manually annotated. Full details on training sets are in supplementary material.

\noindent{$iv$}) {\bf Semi-supervised learning evaluation.} Both active and semi-supervised methods learn from limited labeled data. However, semi-supervised ones assume an equal number of examples per class are labeled, which is not practical in some applications. In contrast, active learning does not use any label information. Therefore, we cannot compare these two learning methods directly. Since selected samples by $K$-means cover a large proportion of all categories, 
% provides a large coverage on the categories, 
one can train semi-supervised learning methods on the small annotated set selected by $K$-means. We follow the procedure of FixMatch \cite{NEURIPS2020_06964dce}, a state-of-the-art semi-supervised method, to evaluate $K$-means, Random and Uniform on CIFAR-10, Flowers-102 \cite{nilsback2008automated}, and DTD-47 \cite{cimpoi2014describing}. % when the labeled set is generated {\em with} or {\em without} annotation information. 

%To show that active methods can exploit the promise of semi-supervised learning, we follow the procedure of FixMatch \cite{NEURIPS2020_06964dce}, a competitive semi-supervised baseline, to evaluate it on CIFAR-10, Flowers-102 \cite{nilsback2008automated}, and DTD-47 \cite{cimpoi2014describing} when the labeled set is generated {\em with} or {\em without} annotation information. 

%%%%%%%%%%%%%
\vspace{-.1in}
\subsection{Performance Analysis on CIFAR-10/100}
Tables \ref{tab:lin_nn_cifar10} through \ref{tab:lin_nn_cifar100} show the performance and category coverage of two {$K$}-means strategies compared to other active learning baselines on CIFAR-10/100. Max-Entropy, Core-set, and VAAL start from randomly selected $0.02\%$ of CIFAR-10 and $0.2\%$ of CIFAR-100. Multi $K$-means starts from $K$-means selected $0.02\%$ of CIFAR-10 and $0.2\%$ of CIFAR-100. We can observe in Table \ref{tab:lin_nn_cifar10} that on CIFAR-10 both {$K$}-means and multi $K$-means outperform all sampling methods in nearest neighbor classification. In linear classification, despite not having equal number of examples per category, {$K$}-means performs better than Uniform at $0.02\%$, $0.04\%$, and $0.6\%$. Tables \ref{tab:cat_cov_cifar100} and \ref{tab:lin_nn_cifar100} compare different selection strategies on CIFAR-100 and show that in both evaluation metrics, $K$-means strategies outperform non-uniform methods and are on par with Uniform.

\vspace{-.1in}
\subsection{Performance Analysis on ImageNet}
\label{sec:imagenet}
Tables \ref{tab:lin_nn_imagenet} and \ref{tab:cat_cov_imagenet} show the scalability of {$K$}-means sampling on the large-scale ImageNet dataset. $0.08\%$ randomly and $K$-means selected ImageNet are used as the initial pools of iterative methods and multi {$K$}-means, respectively. In Table \ref{tab:lin_nn_imagenet}, {$K$}-means outperforms other baselines and has competitive results with Uniform sampling in both evaluation metrics.

Figure \ref{fig:cat_dist} illustrates the distribution of ImageNet categories over their number of occurrences in $0.2\%$ of the unlabeled training data. They are $3,000$ total samples that equals to $3$ per category in average. We prefer the distribution to have a peak at $3$ and be zero otherwise which will happen with Uniform sampling. We see that $K$-means is closer to this peaky distribution compared to Random. Interestingly, Random sampling has almost $4$ times more categories that are not represented in the samples (i.e., zero samples).

\vspace{-.1in}
\subsection{Performance Analysis on ImageNet-LT}
\label{imagenet-lt}

It is important to use a sampling method that does not fail in real-world applications, in which data is not balanced. In Tables \ref{tab:lin_nn_imagenet_lt} and \ref{tab:cat_cov_imagenet_lt}, we use the same frozen backbone that is pre-trained on unlabeled
ImageNet as the feature extractor to compare different sampling strategies on ImageNet-LT.
% We pre-train ResNet-50 using BYOL SSL method \cite{byol} on ImageNet-LT \cite{imagenet_lt}, instead of ImageNet, as a realistic feature extractor and compare it with different sampling strategies in Tables \ref{tab:lin_nn_imagenet_lt} and \ref{tab:cat_cov_imagenet_lt}.
Initial pools of iterative methods and multi $K$-means are $0.8\%$ randomly and $K$-means selected ImageNet-LT, respectively. Uniform sampling chooses equal number of examples per class as long as it does not surpass the class size. Table \ref{tab:lin_nn_imagenet_lt} shows that $K$-means strategies are strong active learning methods in both evaluation metrics with no annotation information. Also, in the appendix \ref{imagenet_lt_appendix}, we report top-1 linear and nearest neighbor classification results of different strategies on ImageNet-LT using an ImageNet-LT pre-trained backbone, as a realistic feature extractor to show that $K$-means is insensitive to category distribution of unlabeled training set.

\begin{table}
\vspace{-.3in}
    \caption{{\bf Top-1 linear (LIN) and nearest neighbor (NN) classification results of different strategies on CIFAR-10.} Both $K$-means strategies outperform all methods in nearest neighbor classification. In linear classification, $K$-means outperforms Max-Entropy, Core-set, VAAL, and Random, and is on par with Uniform. Although in contrast to Uniform, $K$-means does not have equal number of examples per class, this strategy performs better than Uniform in 0.02\%, 0.04\%, and 0.6\% budgets. In both evaluation benchmarks, $K$-means is consistently better than multi $K$-means.}
    \renewcommand{\arraystretch}{0.9}
    % \fontsize{8.0pt}{10.0pt}\selectfont
    \centering
    % \resizebox{\linewidth}{!}{
    \setlength{\tabcolsep}{0.3em}\scalebox{0.81}{
    \begin{tabular}{lccccccccc} \toprule
        &&& \multicolumn{7}{c}{ \textbf{Budgets}} \\ \cmidrule{4-10}
        \textbf{Method} & \textbf{LIN} & \textbf{NN} & \makecell{\texttt{10} \\ 0.02\%} & \makecell{\texttt{20} \\ 0.04\%} & \makecell{\texttt{50} \\ 0.1\%} & \makecell{\texttt{70} \\ 0.14\%} & \makecell{\texttt{100} \\ 0.2\%} & \makecell{\texttt{200} \\ 0.4\%} & \makecell{\texttt{300} \\ 0.6\%} \\ \midrule
        Uniform & \checkmark && $20.4 \pm.2$ & $28.6\pm.0$ &  $\textbf{39.9}\pm.1$ & $\textbf{43.9}\pm.1$ & $\textbf{44.9}\pm.1$ & $\textbf{48.9}\pm.0$ & $51.9\pm.0$\\
\hline
        Random & \checkmark && $21.6\pm.3$ & $28.7\pm.1$ & $32.8\pm.1$ & $36.2\pm.3$ & $43.8\pm.1$ & $48.1\pm.0$ & $50.6\pm.1$ \\
        Max-Entropy & \checkmark && $21.6\pm.3$ & $26.4\pm.1$ & $34.5\pm.2$ & $38.6\pm.0$ & $40.3\pm.1$ & $44.6\pm.1$ & $47.7\pm.0$\\
        Core-set & \checkmark && $21.6\pm.3$ & $27.6\pm.2$ & $34.9\pm.3$ & $35.1\pm.1$ & $38.8\pm.0$ & $43.2\pm.0$ & $46.5\pm.0$\\
        VAAL & \checkmark && $21.6\pm.3$ & $26.4\pm.1$ & $34.7\pm.0$ & $38.8\pm.0$ & $40.8\pm.1$ & $44.6\pm.1$ & $47.7\pm.0$\\
        \rowcolor{lightgray!30} Multi {$K$}-means & \checkmark &&  $\textbf{28.5}\pm.2$ & $28.6\pm.0$ & $35.9\pm.1$ & $39.2\pm.2$ & $41.5\pm.0$ & $48.1\pm.0$ & $51.5\pm.1$ \\
        \rowcolor{lightgray!60} {$K$}-means & \checkmark &&  $\textbf{28.5}\pm.2$ & $\textbf{33.3}\pm.2$ & $\underline{37.6}\pm.1$ & $\underline{43.7}\pm.1$ & $\underline{44.1}\pm.0$ & $\underline{48.8}\pm.2$ &  $\textbf{52.1}\pm.2$ \\ \midrule
        Uniform && \checkmark & $19.8\pm.3$ & $25.7\pm.1$ & $32.2\pm.1$ & $32.9\pm.2$ & $34.2\pm.0$ & $36.4\pm.0$ & $38.3\pm.1$ \\
\hline
        Random && \checkmark & $23.7\pm.2$ & $28.5\pm.1$ & $29.4\pm.2$ & $29.7\pm.1$ & $32.5\pm.1$ & $35.5\pm.0$ & $37.5\pm.0$\\
        Max-Entropy && \checkmark & $23.7\pm.2$ & $24.5\pm.1$ & $28.1\pm.0$ & $29.1\pm.0$ & $27.8\pm.1$ & $28.6\pm.1$ & $30.1\pm.0$ \\
        Core-set && \checkmark & $23.7\pm.2$ & $25.3\pm.1$ & $26.6\pm.1$ & $27.7\pm.0$ & $29.5\pm.0$ & $30.4\pm.1$ & $32.6\pm.0$ \\
        VAAL && \checkmark & $23.7\pm.2$ & $25.1\pm.1$ & $29.1\pm.1$ & $29.8\pm.0$ & $30.1\pm.1$ & $34.5\pm.1$ & $35.8\pm.1$ \\
       \rowcolor{lightgray!30} Multi {$K$}-means && \checkmark & $\textbf{29.5}\pm.1$ & $\underline{29.7}\pm.0$ & $\underline{34.2}\pm.1$ & $\underline{35.1}\pm.0$ & $\underline{36.1}\pm.2$ & $\underline{39.4}\pm.1$ & $\underline{40.2}\pm.1$ \\
        \rowcolor{lightgray!60} {$K$}-means && \checkmark & $\textbf{29.5}\pm.1$ & $\textbf{33.4}\pm.2$ & $\textbf{34.7}\pm.1$  & $\textbf{38.3}\pm.0$ & $\textbf{38.9}\pm.1$ & $\textbf{40.6}\pm.1$ & $\textbf{42.4}\pm.0$ \\ \bottomrule
    \end{tabular}}
    \label{tab:lin_nn_cifar10}
    \vspace{-.3in}
\end{table}

\begin{table}
\vspace{-.3in}
\begin{minipage}[b]{0.47\linewidth}
    \caption{{\bf CIFAR-10 category coverage of selected examples.} $0.02\%$ $K$-means selected instances cover $80\%$ of all CIFAR-10 categories.}
    \vspace{.3in}
    \renewcommand{\arraystretch}{0.7}
    % \fontsize{7.0pt}{10.0pt}\selectfont
    \centering
    \resizebox{\linewidth}{!}{
    \begin{tabular}{@{}lccc@{}} \toprule
        \textbf{Budgets} & \makecell{\texttt{10} \\0.02\%} & \makecell{\texttt{20} \\0.04\%} & \makecell{\texttt{50} \\$\geq$ 0.1\%} \\ \midrule
        Uniform & $\textbf{100}\pm0$ & $\textbf{100}\pm0$ & $100\pm0$  \\
\hline
        Random & $56.7\pm4.7$ & $86.7\pm4.7$  & $100\pm0$ \\
        Max-Entropy & $56.7\pm4.7$ & $68.0\pm2.0$ &   $100\pm0$ \\
        Core-set & $56.7\pm4.7$ & $86.7\pm4.7$  &  $100\pm0$ \\
        VAAL & $56.7\pm4.7$ & $71.0\pm1.0$ & $100\pm0$ \\
        \rowcolor{lightgray!30} Multi {$K$}-means & $80.0\pm0.0$ & $\textbf{100}\pm0.0$ & $100\pm0$  \\
        \rowcolor{lightgray!60} {$K$}-means & $80.0\pm0.0$ & $90.0\pm0.0$  & $100\pm0$ \\ \bottomrule
    \end{tabular}}
    
    \label{tab:cat_cov_cifar10}
\end{minipage}
\hspace*{\fill}
\begin{minipage}[b]{0.47\linewidth}
    \caption{{\bf CIFAR-100 category coverage of selected examples.} After Uniform sampling, both $K$-means methods have a better category coverage than iterative methods and are on par with Random.}
    % \renewcommand{\arraystretch}{0.97}
    % \fontsize{9.0pt}{10.0pt}\selectfont
    \centering
    \resizebox{\linewidth}{!}{
    \begin{tabular}{@{}lcccc@{}} \toprule
        \textbf{Budgets} & \makecell{\texttt{100} \\0.2\%} & \makecell{\texttt{300} \\0.6\%} & \makecell{\texttt{500} \\1\%} & \makecell{\texttt{1000} \\ $\geq$ 2\%}  \\ \midrule
        Uniform & $\textbf{100}\pm0$ & $\textbf{100}\pm0$ & $\textbf{100}\pm0$ & $100\pm0$ \\
\hline
        Random & $60.9\pm3.2$ & $96.7\pm.9$ & $100\pm0$ & $100\pm0$  \\
        Max-Entropy & $60.9\pm3.2$ & $87.7\pm.9$ & $97.0\pm1.4$ & $100\pm0$  \\
        Core-set & $60.9\pm3.2$ & $89.0\pm.0$ &  $97.3\pm1.2$ & $100\pm0$  \\
        VAAL & $60.9\pm3.2$ & $86.3\pm.9$ & $95.7\pm0.9$ & $100\pm0$ \\
        \rowcolor{lightgray!30} Multi {$K$}-means & $68.0\pm0.0$ & $93.0\pm.0$ & $98.0\pm0.0$ & $100\pm0$ \\
        \rowcolor{lightgray!60} {$K$}-means & $68.0\pm0.0$ & $95.0\pm.0$ & $100\pm0$ & $100\pm0$   \\ \bottomrule
    \end{tabular}}
    
    \label{tab:cat_cov_cifar100}
\end{minipage}
\vspace{-.1in}
\end{table}

\begin{table}
    \caption{{\bf Top-1 linear (LIN) and nearest neighbor (NN) classification results of different strategies on CIFAR-100.} In both evaluation benchmarks, $K$-means strategies outperform Max-Entropy, Core-set, VAAL, and Random and are on par with Uniform sampling. Despite having equal number of examples per category, Uniform outperforms $K$-means in linear classification only on budgets larger than $4\%$. $K$-means and multi $K$-means are competitive on CIFAR-100.}
    \renewcommand{\arraystretch}{0.9}
    % \fontsize{9.0pt}{10.0pt}
    \centering
    \resizebox{\linewidth}{!}{
    \begin{tabular}{@{}lccccccccccc@{}} \toprule
        &&& \multicolumn{9}{c}{\bf Budgets} \\ \cline{4-12}
        \textbf{Method} & \textbf{LIN} & \textbf{NN} & \makecell{\texttt{100} \\ 0.2\%} & \makecell{\texttt{300} \\0.6\% } & \makecell{\texttt{500} \\ 1\%} & \makecell{\texttt{1000} \\ 2\% } & \makecell{\texttt{2000} \\ 4\% } & \makecell{\texttt{2500} \\ 5\%} & \makecell{\texttt{4000} \\ 8\%} & 
        \makecell{\texttt{5000} \\ 10\%} &
        \makecell{\texttt{7500} \\ 15\%} \\ \midrule
        Uniform & \checkmark && $10.2\pm.2$ & $18.7\pm.1$ & $21.3\pm.1$ & $27.2\pm.0$ & $\textbf{29.9}\pm.1$ & $30.9\pm.1$ & $\textbf{32.7}\pm.0$ & $\textbf{32.9}\pm.0$ & $\textbf{33.6}\pm.0$\\
\hline
        Random & \checkmark && $10.4\pm.3$ & $16.5\pm.1$ & $20.7\pm.1$ & $24.6\pm.4$ & $29.3\pm.1$ & $29.5\pm.5$ & $30.8\pm.1$ & $31.7\pm.1$ & $32.8\pm.1$\\
        Max-Entropy & \checkmark && $10.4\pm.3$ & $14.6\pm.1$ & $17.2\pm.0$ & $20.8\pm.1$ & $21.9\pm.2$ & $24.6\pm.1$ & $26.1\pm.0$ & $27.6\pm.0$ & $28.8\pm.0$\\
        Core-set & \checkmark && $10.4\pm.3$ & $15.1\pm.0$ & $17.4\pm.1$ & $22.2\pm.1$ & $25.9\pm.0$ & $26.9\pm.0$ & $27.8\pm.1$ & $28.6\pm.0$ & $29.3\pm.1$ \\
        VAAL & \checkmark && $10.4\pm.3$ & $15.9\pm.1$ & $19.1\pm.0$ & $24.1\pm.0$ & $28.4\pm.1$ & $29.9\pm.1$ & $30.9\pm.0$ & $31.6\pm.1$ & $33.1\pm.0$ \\
        \rowcolor{lightgray!30} Multi {$K$}-means & \checkmark && $\textbf{13.4}\pm.1$ & $\textbf{20.0}\pm.1$ & $\underline{22.2}\pm.0$ & $26.1\pm.0$ & $\underline{29.5}\pm.0$ & $30.2\pm.0$ & $31.5\pm.0$ & $31.5\pm.0$ & $32.8\pm.0$ \\
        \rowcolor{lightgray!60} {$K$}-means & \checkmark && $\textbf{13.4}\pm.1$ & $\underline{18.8}\pm.1$ & $\textbf{23.7}\pm.1$ & $\textbf{27.1}\pm.0$ & $29.4\pm.0$ & $\textbf{31.1}\pm.1$ & $\underline{32.5}\pm.0$ & $\underline{32.8}\pm.1$ & $\underline{33.4}\pm.0$\\ \midrule
        Uniform && \checkmark & $10.1\pm.3$ & $13.8\pm.1$ & $15.3\pm.1$ & $17.8\pm.0$ & $20.2\pm.0$ &
        $21.9\pm.0$ & $24.1\pm.1$ & $\textbf{25.8}\pm.1$ & $\textbf{30.0}\pm.1$\\
\hline
        Random && \checkmark & $8.3\pm.1$ & $12.8\pm.1$ & $15.0\pm.2$ & $16.9\pm.1$ & $19.7\pm.1$ & $20.8\pm.0$ & $22.0\pm.1$ & $23.2\pm.0$ & $25.5\pm.0$\\
        Max-Entropy && \checkmark & $8.3\pm.1$ & $10.1\pm.0$ & $10.9\pm.0$ & $12.1\pm.1$ & $12.5\pm.1$ & $12.7\pm.1$ & $12.9\pm.0$ & $13.1\pm.0$ & $13.6\pm.1$\\
        Core-set && \checkmark & $8.3\pm.1$ & $10.4\pm.1$ & $10.9\pm.0$ & $13.3\pm.0$ & $16.4\pm.1$ & $16.8\pm.0$ & $18.2\pm.0$ & $18.9\pm.0$ & $20.7\pm.1$\\
        VAAL && \checkmark & $8.3\pm.1$ & $12.1\pm.0$ & $13.7\pm.1$ & $16.7\pm.0$ & $19.2\pm.1$ & $20.4\pm.0$ & $22.1\pm.1$ & $23.1\pm.0$ & $24.9\pm.0$ \\
        \rowcolor{lightgray!30} Multi {$K$}-means && \checkmark & $\textbf{13.7}\pm.2$ & $\textbf{17.2}\pm.0$ & $\underline{18.0}\pm.1$ & $\underline{20.4}\pm.1$ & $\underline{22.9}\pm.0$ & $\underline{23.5}\pm.0$ & $\underline{24.2}\pm.1$ & $\underline{25.1}\pm.0$ & $\underline{27.2}\pm.0$\\
        \rowcolor{lightgray!60} {$K$}-means && \checkmark & $\textbf{13.7}\pm.2$ & $\underline{17.1}\pm.0$ & $\textbf{19.4}\pm.1$ & $\textbf{21.1}\pm.1$ & $\textbf{23.1}\pm.0$ &
        $\textbf{23.6}\pm.1$ & $\textbf{24.5}\pm.0$ & $24.9\pm.1$ & $26.1\pm.1$\\ \bottomrule
    \end{tabular}}
    \label{tab:lin_nn_cifar100}
\vspace{-.3in}
\end{table}

%%%%%%%%%%%%

\begin{table}
\vspace{-.1in}
    \caption{{\bf Top-1 linear (LIN) and nearest neighbor (NN) classification results of different strategies on ImageNet.} $K$-means outperforms all non-uniform sampling methods. Linear classification results of Uniform sampling that takes advantage of equal number of examples per class are better than $K$-means only at $5\%$ and $10\%$ by a small margin.}
    \renewcommand{\arraystretch}{0.9}
    % \fontsize{9.0pt}{10.0pt}\selectfont
    \centering
    \resizebox{\linewidth}{!}{
    \begin{tabular}{@{}lcccccccccc@{}} \toprule
        &&& \multicolumn{8}{c}{\bf Budgets} \\ \cline{4-11}
        \textbf{Method} & \textbf{LIN} & \textbf{NN} & \makecell{\texttt{1K} \\0.08\%} & \makecell{\texttt{3K} \\0.2\%} & \makecell{\texttt{7K} \\0.5\%} & \makecell{\texttt{13K} \\1\%} & \makecell{\texttt{26K} \\2\%} & \makecell{\texttt{64K} \\5\%} & \makecell{\texttt{128K} \\10\%} & \makecell{\texttt{192K} \\15\%} \\ \midrule
        Uniform & \checkmark && $19.2\pm.3$ & $31.9\pm.3$ & $41.0\pm.3$ & $46.0\pm.1$ & $49.9\pm.0$ & $\textbf{54.2}\pm.1$ & $\textbf{56.7}\pm.1$ & $57.9\pm.1$ \\
\hline
        Random & \checkmark && $15.8\pm.0$ & $28.0\pm.4$ & $39.2\pm.3$ & $45.1\pm.1$ & $49.7\pm.1$ & $\underline{54.0}\pm.1$ & $56.6\pm.1$ & $57.9\pm.1$ \\
        Max-Entropy & \checkmark && $15.8\pm.0$ & $19.4\pm.0$ & $25.6\pm.0$ & $33.7\pm.0$ & $41.3\pm.0$ & $48.9\pm.1$ & $51.9\pm.1$ & $54.3\pm.1$ \\
        Core-set & \checkmark && $15.8\pm.0$ & $25.6\pm.1$ & $33.3\pm.0$ & $39.6\pm.0$ & $45.7\pm.1$ & $51.3\pm.0$ & $54.9\pm.1$ & $56.6\pm.1$ \\
        VAAL & \checkmark && $15.8\pm.0$ & $27.7\pm.1$ & $34.9\pm.2$ & $42.8\pm.1$ & $49.2\pm.2$ & $53.6\pm.1$ & $56.0\pm.1$ & $57.4\pm.0$ \\
        \rowcolor{lightgray!30} Multi {$K$}-means & \checkmark && $\textbf{24.6}\pm.0$ & $\underline{34.1}\pm.1$ & $\underline{41.1}\pm.0$ & $45.3\pm.0$ & $49.5\pm.0$ & $53.9\pm.0$ & $56.3\pm.0$ & $57.5\pm.1$  \\
        \rowcolor{lightgray!60} {$K$}-means & \checkmark && $\textbf{24.6}\pm.0$ & $\textbf{35.7}\pm.0$ & $\textbf{42.6}\pm.1$ & $\textbf{46.9}\pm.1$ & $\textbf{50.7}\pm.0$ & $\underline{54.0}\pm.1$ & $\underline{56.6}\pm.0$ & $\textbf{58.0}\pm.1$  \\ \midrule
        Uniform && \checkmark & $29.5\pm.1$ & $35.7\pm.1$ & $38.9\pm.2$ & $41.1\pm.1$ & $43.2\pm.0$ & $45.6\pm.0$ & $47.6\pm.2$ & $48.6\pm.0$\\
\hline
        Random && \checkmark & $22.8\pm.2$ & $33.2\pm.7$ & $38.4\pm.2$ & $40.8\pm.0$ & $42.2\pm.1$ & $45.4\pm.1$ & $47.3\pm.1$ & $48.3\pm.1$\\
        Max-Entropy && \checkmark & $22.8\pm.2$ & $24.5\pm.1$ & $27.2\pm.1$ & $30.3\pm.0$ & $33.3\pm.1$ & $36.2\pm.1$ & $37.6\pm.0$ & $38.6\pm.0$\\
        Core-set && \checkmark & $22.8\pm.2$ & $30.7\pm.0$ & $34.8\pm.1$ & $37.5\pm.1$ & $39.7\pm.1$ & $42.0\pm.1$ & $43.7\pm.2$ & $44.6\pm.1$ \\
        VAAL && \checkmark & $22.8\pm.2$ & $32.8\pm.1$ & $36.2\pm.0$ & $39.7\pm.1$ & $42.6\pm.0$ & $45.3\pm.0$ & $46.7\pm.1$ & $47.9\pm.0$ \\
        \rowcolor{lightgray!30} Multi {$K$}-means && \checkmark & $\textbf{31.6}\pm.1$ & $\underline{38.2}\pm.0$ & $\underline{41.4}\pm.0$ & $\underline{43.3}\pm.0$ & $\underline{45.2}\pm.0$ & $\underline{47.2}\pm.0$ & $\textbf{48.6}\pm.0$ & $\textbf{49.4}\pm.0$ \\
        \rowcolor{lightgray!60} {$K$}-means && \checkmark & $\textbf{31.6}\pm.1$ & $\textbf{39.9}\pm.0$ & $\textbf{42.7}\pm.0$ & $\textbf{44.0}\pm.1$ & $\textbf{45.5}\pm.0$ & $\textbf{46.8}\pm.1$ & $\underline{48.1}\pm.1$ & $\underline{48.8}\pm.0$ \\ \bottomrule
    \end{tabular}}
    
    \label{tab:lin_nn_imagenet}
\end{table}

\begin{table}
\caption{{\bf Category distribution of sampled ImageNet.} (a) $K$-means covers almost $98\%$ of all classes with only $3,000$ examples while iterative methods require at least $7,000$ examples to reach this coverage; (b) Compared to Random selection, $K$-means has a sharper distribution around $3$ which means selected images are distributed more uniformly across the categories.}
% \vspace{-.1in}
\begin{minipage}{0.47\linewidth}
\vspace{-.25in}
    \subcaption{{\bf ImageNet category coverage of selected examples.}}
    % \vspace{.35in}
    % \renewcommand{\arraystretch}{1.1}
    % \fontsize{9.0pt}{10.0pt}\selectfont
    \centering
    \resizebox{\textwidth}{!}{
    \begin{tabular}{lcccc} \toprule
        \textbf{Budgets} & \makecell{\texttt{1K} \\0.08\%} & \makecell{\texttt{3K} \\0.2\%} & \makecell{\texttt{7K} \\0.5\%} & \makecell{\texttt{13K} \\$\geq$ 1\%} \\ \midrule
        Uniform & $\textbf{100}\pm0$  & $\textbf{100}\pm0$ & $\textbf{100}\pm0$ & $\textbf{100}\pm0$ \\
\hline
        Random & $62.9\pm.2$ & $94.6\pm.4$  & $100\pm0$ & $100\pm0$  \\
        Max-Entropy & $62.9\pm.2$ & $84.3\pm.5$ & $94.8\pm.2$ & $100\pm0$ \\
        Core-set &  $62.9\pm.2$ & $87.9\pm.1$  & $97.0\pm.5$ & $100\pm0$  \\
        VAAL & $62.9\pm.2$ & $94.6\pm.1$ & $98.1\pm.3$ & $100\pm0$  \\
        \rowcolor{lightgray!30} Multi {$K$}-means & $72.2\pm.1$ & $97.0\pm.0$ & $99.8\pm.0$ & $100\pm0$  \\ 
        \rowcolor{lightgray!60} {$K$}-means & $72.2\pm.1$ & $97.8\pm.2$ & $99.9\pm.0$ & $100\pm0$  \\ \bottomrule
    \end{tabular}}
    % \vspace{.3in}
    \label{tab:cat_cov_imagenet}
\end{minipage}
\hspace*{\fill}
\begin{minipage}{0.47\linewidth}
    % \begin{figure}
    \subcaption{{\bf Category distribution of $\textbf{3,000}$ samples of ImageNet ($\textbf{0.2\%}$).}}
      \vspace{-.1in}
    \begin{center}
    
      \includegraphics[width=.9\textwidth]{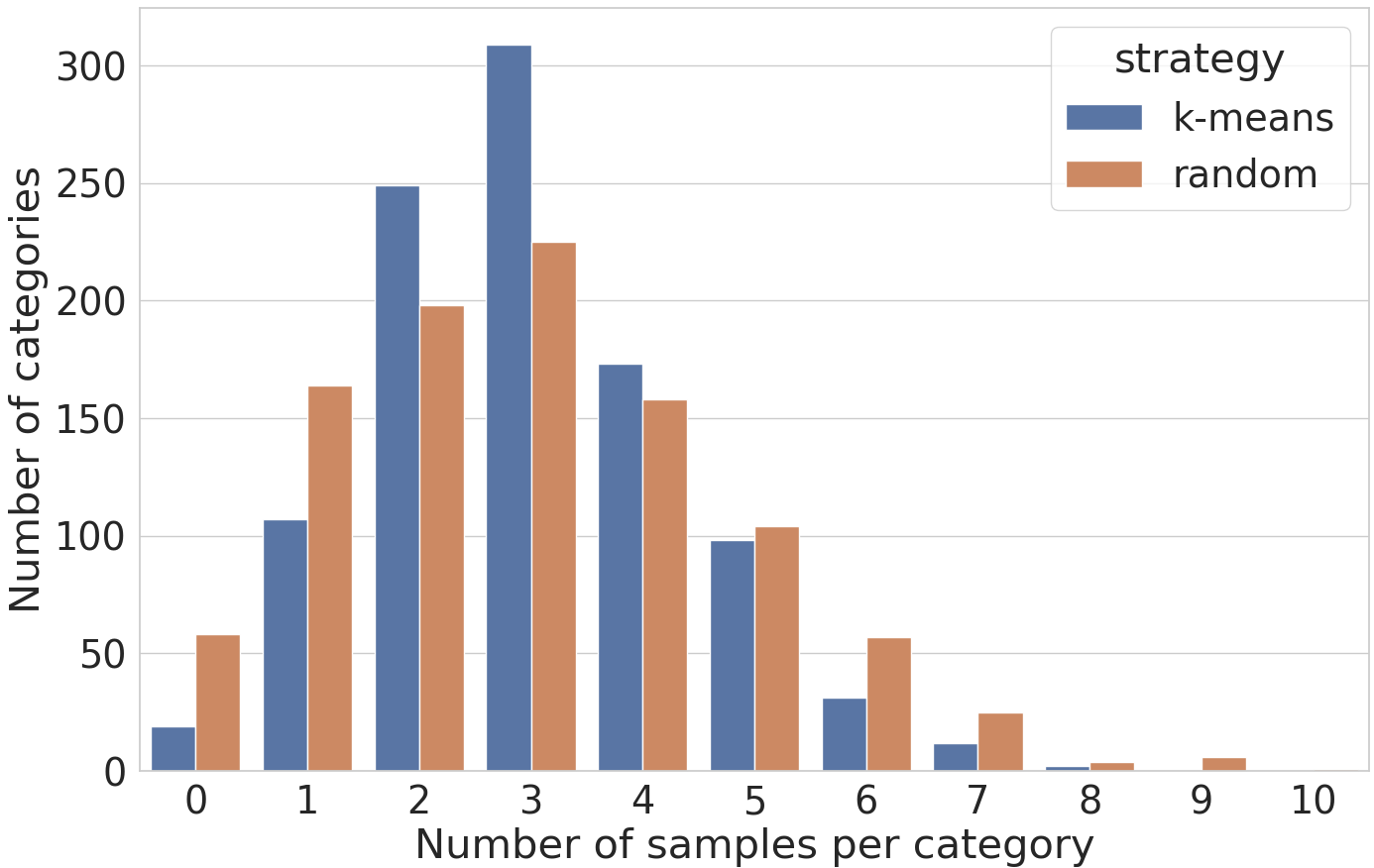}
    \end{center}
      
    \label{fig:cat_dist}
    % \end{figure}
\end{minipage}
% \vspace{-.4in}
\end{table}

\begin{table}
\vspace{-.4in}
    \caption{{\bf Top-1 linear (LIN) and nearest neighbor (NN) classification results of different strategies on ImageNet-LT.} We use the same frozen backbone pre-trained on unlabeled ImageNet as the feature extractor.
    % ImageNet-LT pre-trained ResNet-50 as a more realistic feature extractor. 
    With no labels information, $K$-means is a strong selection baseline in low budgets in both linear and nearest neighbor classification compared to prior works.}
    \renewcommand{\arraystretch}{0.85}
    % \fontsize{8.0pt}{5.0pt}\selectfont
    \centering
    % \resizebox{\linewidth}{!}{
    \setlength{\tabcolsep}{0.3em}\scalebox{0.81}{
    \begin{tabular}{@{}lcccccccc@{}} \toprule
        && & \multicolumn{6}{c}{\bf Budgets} \\ \cline{4-9}
        \textbf{Method} & \textbf{LIN} & \textbf{NN} & \makecell{\texttt{1K} \\0.8\%} & \makecell{\texttt{3K} \\3\%} & \makecell{\texttt{5K} \\4\%} & \makecell{\texttt{7K} \\6\%} & \makecell{\texttt{9K} \\8\%} &
        \makecell{\texttt{12K} \\10\%} \\ \midrule
        Uniform & \checkmark && $\textbf{19.4}\pm.1$ & $\textbf{31.8}\pm.1$ & $\textbf{37.2}\pm.2$ & $\textbf{40.5}\pm.0$ & $\textbf{42.7}\pm.1$ & $\textbf{44.8}\pm.0$\\
\hline
        Random & \checkmark && $13.0\pm.2$ & $21.6\pm.1$ & $26.7\pm.1$ & $29.8\pm.0$ & $32.1\pm.0$ & $34.3\pm.0$\\
        Max-Entropy & \checkmark && $13.0\pm.2$ & $16.3\pm.0$ & $19.8\pm.1$ & $22.9\pm.0$ & $25.6\pm.1$ & $28.6\pm.1$ \\
        Core-set & \checkmark && $13.0\pm.2$ & $22.7\pm.0$ & $26.9\pm.0$ & $30.3\pm.1$ & $32.1\pm.0$ & $34.9\pm.0$ \\
        VAAL & \checkmark && $13.0\pm.2$ & $21.8\pm.0$ & $25.9\pm.0$ & $28.7\pm.1$ & $30.8\pm.0$ & $33.4\pm.0$ \\
        \rowcolor{lightgray!30} Multi {$K$}-means & \checkmark && $\underline{18.1}\pm.0$ & $24.7\pm.0$ & $27.4\pm.1$ & $29.4\pm.0$ & $30.8\pm.0$ & $33.5\pm.0$ \\ 
        \rowcolor{lightgray!60} {$K$}-means & \checkmark && $\underline{18.1}\pm.0$ & $\underline{25.9}\pm.1$ & $\underline{29.4}\pm.0$ & $\underline{31.7}\pm.2$ & $\underline{33.6}\pm.0$ & $\underline{35.7}\pm.1$  \\ 
\midrule
        Uniform && \checkmark & $\textbf{29.2}\pm.0$ & $\textbf{35.9}\pm.1$ & $\textbf{37.9}\pm.1$ & $\textbf{38.8}\pm.1$ & $\textbf{39.5}\pm.0$ & $\textbf{40.4}\pm.0$\\
\hline
        Random && \checkmark & $19.3\pm.1$ & $27.6\pm.0$ & $31.3\pm.0$ & $33.1\pm.0$ & $34.4\pm.0$ & $35.7\pm.0$ \\
        Max-Entropy && \checkmark & $19.3\pm.1$ & $22.3\pm.1$ & $24.2\pm.1$ & $25.9\pm.0$ & $27.0\pm.0$ & $28.8\pm.0$ \\
        Core-set && \checkmark & $19.3\pm.1$ & $29.3\pm.0$ & $31.9\pm.0$ & $33.7\pm.0$ & $34.6\pm.0$ & $35.9\pm.0$ \\
        VAAL && \checkmark & $19.3\pm.1$ & $26.9\pm.0$ & $30.4\pm.0$ & $32.2\pm.0$ & $33.4\pm.1$ & $34.7\pm.0$ \\
        \rowcolor{lightgray!30} Multi {$K$}-means && \checkmark & $\underline{24.3}\pm.0$ & $29.6\pm.0$ & $31.2\pm.0$ & $32.2\pm.1$ & $32.9\pm.1$ & $34.9\pm.0$ \\
        \rowcolor{lightgray!60} {$K$}-means && \checkmark & $\underline{24.3}\pm.0$ & $\underline{31.5}\pm.0$ & $\underline{34.3}\pm.1$ & $\underline{35.3}\pm.0$ & $\underline{36.3}\pm.0$ & $\underline{36.9}\pm.0$  \\ \bottomrule
    \end{tabular}
}
\label{tab:lin_nn_imagenet_lt}
\vspace{-.3in}
\end{table}

\begin{table}
    \caption{{\bf ImageNet-LT category coverage of selected examples.} Uniform selects equal number of samples per class as long as that class contains examples.}
    % \renewcommand{\arraystretch}{0.7}
    % \fontsize{13.0pt}{10.0pt}\selectfont
    \centering
    \setlength{\tabcolsep}{0.3em}\scalebox{0.81}{
    \begin{tabular}{@{}lcccccc@{}} \toprule
        \textbf{Budgets} & \makecell{\texttt{1K} \\0.8\%} & \makecell{\texttt{3K} \\3\%} & \makecell{\texttt{5K} \\4\%} & \makecell{\texttt{7K} \\6\%} & \makecell{\texttt{9K} \\8\%} &
        \makecell{\texttt{12K} \\10\%} \\ \midrule
        Uniform & $\textbf{100}\pm0$  & $\textbf{100}\pm0$ & $\textbf{100}\pm0$ & $\textbf{100}\pm0$ & $\textbf{100}\pm0$ & $\textbf{100}\pm0$ \\
\hline
        Random & $48.6\pm.3$ & $76.1\pm1.0$ & $84.9\pm.5$ & $89.6\pm.3$ & $92.2\pm.4$ & $94.8\pm.3$  \\
        Max-Entropy
        & $48.6\pm.3$ & $69.6\pm.2$ & $78.8\pm.1$ & $84.2\pm.1$ & $89.3\pm.0$ & $92.0\pm.0$ \\
        Core-set & $48.6\pm.3$ & $77.5\pm.4$ & $86.7\pm.4$ & $91.1\pm.8$ & $93.5\pm.0$ & $95.8\pm.3$  \\
        VAAL & $48.6\pm.3$ & $73.9\pm.3$ & $83.1\pm.1$ & $87.6\pm.1$ & $90.8\pm.1$ & $93.6\pm.1$ \\
        \rowcolor{lightgray!30} Multi {$K$}-means
        & $51.8\pm.0$ & $71.8\pm.0$ & $79.7\pm.0$ & $84.3\pm.0$ & $88.3\pm.0$ & $92.1\pm.0$ \\ 
        \rowcolor{lightgray!60} {$K$}-means & $51.8\pm.0$ & $75.8\pm.0$ & $86.8\pm.0$ & $90.8\pm.0$ & $92.2\pm.0$ & $95.1\pm.0$ \\ \bottomrule
    \end{tabular}}
    
    \label{tab:cat_cov_imagenet_lt}
\vspace{-.1in}
\end{table}

\subsection{Evaluation on Fine-grained Tasks}
\label{fine_grained}
In Table \ref{tab:transfer}, we analyze $K$-means clustering on datasets that are finer-grained compared to ImageNet. This table shows that the linear classification results of $K$-means on fine-grained datasets are consistently better than Random and are on par with Uniform.

\begin{table}
\vspace{-.2in}
    \caption{\textbf{Evaluation on fine-grained tasks.} The same frozen backbone that is pre-trained on unlabeled ImageNet is used to extract feature embeddings. For each dataset, a small subset of training data is annotated and is used to train a linear classifier on top of the frozen backbone. With no label information, $K$-means consistently outperforms Random and is on par with Uniform.}
    % \renewcommand{\arraystretch}{0.7}
    % \fontsize{13.0pt}{10.0pt}\selectfont
    \centering
    \setlength{\tabcolsep}{0.3em}\scalebox{0.81}{
    \begin{tabular}{@{}l c cc c cc c cc@{}} \toprule
         && \multicolumn{2}{c}{\textbf{Flowers-102}} && \multicolumn{2}{c}{\bf DTD-47} && \multicolumn{2}{c}{\bf Aircraft-100} \\ \cmidrule{3-4} \cmidrule{6-7} \cmidrule{9-10}
         \textbf{Methods} && \makecell{\texttt{1$\times$102} \\ 10\%} & \makecell{\texttt{4$\times$102} \\ 40\%} && \makecell{\texttt{1$\times$47} \\ 2.5\%} & \makecell{\texttt{4$\times$47} \\ 10\%} &&
         \makecell{\texttt{1$\times$100} \\ 2\%} &
         \makecell{\texttt{4$\times$100} \\ 7.5\%} 
         \\ \cmidrule{3-4} \cmidrule{6-7} \cmidrule{9-10}
        Uniform && $81.56$ \tiny{$\pm$1.1} & $82.70$ \tiny{$\pm$0.0} && $64.23$ \tiny{$\pm$1.6} & $65.87$ \tiny{$\pm$1.1} && $36.73$ \tiny{$\pm$0.3} & $37.43$ \tiny{$\pm$0.3} \\ \midrule
        Random && $80.56$ \tiny{$\pm$1.6} & $82.06$ \tiny{$\pm$1.3} &&  $53.86$ \tiny{$\pm$0.6} & $61.06$ \tiny{$\pm$0.7} && $36.10$ \tiny{$\pm$1.7} & $37.36$ \tiny{$\pm$0.1} \\
         \rowcolor{lightgray!30}$K$-means && $82.20$ \tiny{$\pm$1.0} & $82.76$ \tiny{$\pm$1.2} && $64.30$ \tiny{$\pm$0.0} & $65.96$ \tiny{$\pm$0.7} && $37.20$ \tiny{$\pm$0.0} & $38.67$ \tiny{$\pm$0.2}   \\ \bottomrule
    \end{tabular}
    }
    \label{tab:transfer}
\vspace{-.1in}
\end{table}

\subsection{Semi-Supervised Learning Evaluation}
\label{semi-sup}
Table \ref{tab:semi_sup} shows semi-supervised evaluation results of FixMatch \cite{NEURIPS2020_06964dce} when the labeled set is selected {\em with} label information, by Uniform and Uniform($K$-means), or {\em without} it, by Random and $K$-means. In this table, Uniform($K$-means) uses number of classes to cluster the dataset and selects equal number of examples per cluster. As shown in Table \ref{tab:semi_sup}, $K$-means outperforms Random over all datasets with no label information. Taking the advantage of annotations, Uniform($K$-means) performs better than Uniform in low budgets.
\begin{table}
\vspace{-.2in}
    \caption{\textbf{Semi-supervised evaluation with FixMatch \cite{NEURIPS2020_06964dce}.} The scores are top-1 accuracies (in \%) of the model on the test set. In contrast to Random and $K$-means, Uniform and Uniform($K$-means) take advantage of annotation information to sample a labeled set. With no labels, $K$-means performs consistently better than Random. Using labels, Uniform($K$-means) outperforms Uniform in low budgets. The results are from 1 repetition of the experiments.}
    \renewcommand{\arraystretch}{0.6}
    % \fontsize{13.0pt}{10.0pt}\selectfont
    \centering
    % \resizebox{\linewidth}{!}{
    \setlength{\tabcolsep}{0.5em}\scalebox{0.81}{
    \begin{tabular}{@{}l ccc c ccc c ccc@{}} \toprule
         & \multicolumn{3}{c}{\textbf{CIFAR-10}} && \multicolumn{3}{c}{\textbf{Flowers-102}} && \multicolumn{3}{c}{\textbf{DTD-47}} \\ \cmidrule{2-4} \cmidrule{6-8} \cmidrule{10-12}
         \textbf{Methods} & \makecell{\texttt{1$\times$10} \\ 0.02\%} & \makecell{\texttt{4$\times$10} \\ 0.08\%} & \makecell{\texttt{10$\times$10} \\ 0.2\%} && \makecell{\texttt{1$\times$102} \\ 10\%} & \makecell{\texttt{3$\times$102} \\ 30\%} & \makecell{\texttt{4$\times$102} \\ 40\%} && \makecell{\texttt{1$\times$47} \\ 2.5\%} & \makecell{\texttt{3$\times$47} \\ 7.5\%} & \makecell{\texttt{4$\times$47} \\ 10\%} \\  \cmidrule{2-4} \cmidrule{6-8} \cmidrule{10-12}
         
         \rowcolor{lightgray!30}& \multicolumn{11}{c}{\textit{With Labels}} \\ \\
         
         Uniform              & $54.79$  & $88.76$ & $\textbf{91.20}$         && $15.22$ & $34.33$ & $\textbf{42.54}$          && $8.88$ & $18.40$ & $\textbf{23.19}$ \\
         Uniform({\em $K$-means}) & $\textbf{57.53}$ & $\textbf{89.11}$ & $89.40$ && $\textbf{18.08}$ & $\textbf{37.79}$ & $41.60$ && $\textbf{16.49}$ & $\textbf{22.29}$ & $22.39$ \\ \cmidrule{2-12}
         
         \rowcolor{lightgray!30}& \multicolumn{11}{c}{\textit{Without Labels}} \\ \\
         
        Random               & $38.69$ & $60.21$ & $86.73$                             && $13.81$ & $31.42$ & $42.51$                          && $8.30$ & $16.86$ & $21.65$ \\
        $K$-means & $\textbf{43.37}$  & $\textbf{84.46}$ & $\textbf{86.96}$ && $\textbf{19.87}$ & $\textbf{37.60}$ & $\textbf{46.09}$ && $\textbf{14.36}$ & $\textbf{19.04}$ & $\textbf{23.56}$ \\ \bottomrule
    \end{tabular}}
    \label{tab:semi_sup}
% \vspace{-.5in}
\end{table}

\section{Ablation Study}

In this section, we perform ablation study on the initial labeled pool and the feature extraction backbone.

\subsection{Effect of Initial Pool}

Interestingly, Random sampling performs better than iterative active learning baselines at both accuracy and category coverage in some experiments. This may happen since a very small random initial labeled pool might not be representative of all categories and cannot be a strong starting point for iterative methods. To examine this hypothesis, we investigate how the strategy of sampling and size of the initial pool affect the performance of iterative methods.

\vspace{-.2in}

\begin{table}
\begin{minipage}[b]{0.45\textwidth}
    \caption{{\bf Effect of a larger ($\textbf{2\%}$) initial labeled set on ImageNet linear classification results.} VAAL and Core-set perform better than $K$-means and Random sampling using $10\%$ of the unlabeled data. Results of Random and $K$-means are repeated from Table \ref{tab:lin_nn_imagenet}.}
    % \renewcommand{\arraystretch}{0.6}
    % \fontsize{9.0pt}{10.0pt}\selectfont
    \centering
    \resizebox{\linewidth}{!}{
    \begin{tabular}{@{}lccc@{}} \toprule
     \textbf{Budget} & \makecell{\texttt{26K} \\2\%} & \makecell{\texttt{64K} \\5\%} & \makecell{\texttt{128K} \\10\%} \\ \midrule
        Max-Entropy & $49.7\pm.1$ & $51.8\pm.0$ & $56.6\pm.0$   \\
        Core-set & $49.7\pm.1$ & $52.5\pm.1$ & $\underline{56.7}\pm.0$  \\
        VAAL & $49.7\pm.1$ & $53.8\pm.1$ & $\textbf{56.8}\pm.1$  \\ \hline
        Random & $49.7\pm.1$ & $\textbf{54.0}\pm.1$ & $56.6\pm.1$ \\
        \rowcolor{lightgray!30} {$K$}-means & $\textbf{50.7}\pm.0$ & $\textbf{54.0}\pm.1$ & $56.6\pm.0$ \\ \bottomrule
    \end{tabular}}
    \label{tab:initial_pool_size}
\end{minipage}
\hspace*{\fill}
\begin{minipage}[b]{0.45\textwidth}
    \caption{{\bf Effect of $\textbf{0.08\%}$ $\textbf{K}$-means selected initial set of ImageNet on linear classification results.} Despite an improvement in the performance of iterative methods, $K$-means still outperforms them in low budgets. Results of $K$-means are repeated from Table \ref{tab:lin_nn_imagenet}.}
    % \vspace{.15in}
    \renewcommand{\arraystretch}{1.2}
    % \fontsize{9.0pt}{10.0pt}\selectfont
    \centering
    \resizebox{\linewidth}{!}{\begin{tabular}{lccc} \toprule
     \textbf{Budget} & \makecell{\texttt{1K} \\0.08\%} & \makecell{\texttt{3K} \\0.2\%} & \makecell{\texttt{7K} \\0.5\%} \\ \midrule
        Max-Entropy & $24.6\pm.0$ & $26.2\pm.1$ & $30.0\pm.0$   \\
        Core-set & $24.6\pm.0$ & $28.2\pm.0$ & $33.6\pm.0$ \\
        VAAL & $24.6\pm.0$ & $28.9\pm.1$ & $35.0\pm.0$ \\ \rowcolor{lightgray!30} {$K$}-means & $24.6\pm.0$ & $\textbf{35.7}\pm.0$ & $\textbf{42.6}\pm.1$ \\ \bottomrule
    \end{tabular}}
    \label{tab:initial_pool_strategy}
\end{minipage}
\vspace{-.2in}
\end{table}

\vspace{.05in}
\noindent{\bf Effect of size.} We repeat the experiments with the same setting as Section \ref{sec:imagenet} for a larger randomly sampled initial set of $2\%$. By comparing linear classification results of Tables \ref{tab:lin_nn_imagenet} and \ref{tab:initial_pool_size}, we find that all three iterative methods perform better using a larger initial pool. As a result, iterative methods are suitable options when large initial pools are available.

\vspace{.05in}
\noindent{\bf Effect of sampling strategy.} We repeat the analyses in Section \ref{sec:imagenet} with {$K$}-means selected instead of randomly selected $0.08\%$ of ImageNet as the initial pool. The linear classification results on $0.2\%$ and $0.5\%$ of ImageNet are shown in Table \ref{tab:initial_pool_strategy}. Comparing Tables \ref{tab:lin_nn_imagenet} and \ref{tab:initial_pool_strategy} demonstrates that although the performance of iterative methods are improved with a richer initialization, there is still a great gap with {$K$}-means in low budgets.

\begin{table*}[t]
    \caption{{\bf Fine-tuning and linear classification results of ablation study on sampling ImageNet using $\textbf{K}$-means.} SSL refers to using SSL pre-trained weights and Rand refers to using randomly initialized weights for backbones. Initializing both selection and classification backbones with SSL pre-trained or random features is an empirical higher and lower bound for our experiments, respectively. In the fine-tuning process, forgetting happens and causes the gap with the linear classification counterpart.}
    % \renewcommand{\arraystretch}{0.5}
    % \fontsize{9.0pt}{10.0pt}\selectfont
    \centering
    \resizebox{\linewidth}{!}{
    \begin{tabular}{ccccccccccccc} \toprule
         \multicolumn{2}{c}{\makecell{\textbf{Selection} \\ \textbf{backbone}}} & \multicolumn{2}{c}{\makecell{\textbf{Classification} \\ \textbf{backbone}}} & \makecell{\textbf{Fine-} \\ \textbf{tuning}} &  \multicolumn{8}{c}{\textbf{Budgets}}  \\ \midrule
        SSL & Rand & SSL & Rand && \makecell{\texttt{1K} \\0.08\%} & \makecell{\texttt{3K} \\0.2\%} & \makecell{\texttt{7K} \\0.5\%} & \makecell{\texttt{13K} \\1\%} & \makecell{\texttt{26K} \\2\%} & \makecell{\texttt{64K} \\5\%} & \makecell{\texttt{128K} \\10\%} & \makecell{\texttt{256K} \\20\%} \\ \midrule
         & \checkmark  &  & \checkmark  & No & $0.38\pm.0$ & $0.75\pm.0$ & $1.07\pm.0$ & $1.41\pm.1$ & $2.05\pm.0$ & $3.46\pm.0$ & $4.26\pm.1$ & $5.08\pm.1$ \\
         & \checkmark  & \checkmark &  & No & $15.1\pm.0$ & $26.5\pm.0$ & $36.6\pm.1$ & $43.8\pm.0$ & $49.1\pm.0$ & $54.0\pm.0$ & $56.6\pm.0$ & $58.7\pm.0$ \\
         \checkmark &  &  & \checkmark & No & $0.47\pm.0$ & $0.87\pm.1$ & $1.26\pm.1$ & $1.79\pm.1$ & $2.37\pm.1$ & $3.35\pm.1$ & $4.19\pm.1$ & $5.01\pm.1$ \\
         \rowcolor{lightgray!30}\checkmark &  & \checkmark &  & No & $\textbf{24.6}\pm.0$ & $\textbf{35.7}\pm.0$ & $\textbf{42.6}\pm.1$ & $\textbf{46.9}\pm.1$ & $\textbf{50.7}\pm.0$ & $\textbf{54.0}\pm.1$ & $\textbf{56.6}\pm.0$ & $\underline{58.9}\pm.0$  \\ \midrule
         & \checkmark  &  & \checkmark  & Yes & $0.73\pm.0$ & $1.50\pm.0$ & $3.49\pm.1$ & $8.43\pm.3$ & $17.7\pm.3$ & $35.3\pm.6$ & $48.4\pm.2$ & $58.3\pm.2$ \\
         & \checkmark  & \checkmark &  & Yes & $12.2\pm.1$ & $22.7\pm.2$ & $32.2\pm.1$ & $38.7\pm.1$ & $44.9\pm.1$ & $51.3\pm.0$ & $55.4\pm.2$ & $59.9\pm.1$\\
         \checkmark &  &  & \checkmark & Yes & $1.22\pm.0$ & $2.43\pm.0$ & $5.63\pm.2$ & $11.4\pm.2$ & $19.9\pm.3$ & $36.2\pm.3$ & $48.9\pm.0$ & $58.4\pm.1$ \\
         \rowcolor{lightgray!30}\checkmark &  & \checkmark &  & Yes & $19.5\pm.2$ & $30.3\pm.2$ & $37.3\pm.1$ & $41.8\pm.2$ & $46.5\pm.2$ & $51.5\pm.2$ & $55.6\pm.1$ & $\textbf{60.0}\pm.1$\\ \bottomrule
    \end{tabular}}
    
    \label{tab:ablation_study}
    \vspace{-.1in}
\end{table*}

\subsection{Ablating Feature Extraction Backbone}

\begin{table}[t]
    \caption{{\bf Effect of MoCo-v2 pre-trained R50 on top-1 linear (LIN) and nearest neighbor (NN) classification results of ImageNet.} The superiority of $K$-means is insensitive to the choice of network architecture and SSL pre-training method. The results are from $1$ repetition of the experiments.}
    \renewcommand{\arraystretch}{0.9}
    \centering
    \setlength{\tabcolsep}{0.5em}\scalebox{0.81}{
    \begin{tabular}{@{}lccccccccc@{}} \toprule
        &&& \multicolumn{6}{c}{\bf Budgets} \\ \cline{4-9}
        \textbf{Method} & \textbf{LIN} & \textbf{NN}  & \makecell{\texttt{1K} \\0.08\%} & \makecell{\texttt{3K} \\0.2\%} & \makecell{\texttt{7K} \\0.5\%} & \makecell{\texttt{13K} \\1\%} & \makecell{\texttt{26K} \\2\%} & \makecell{\texttt{64K} \\5\%} \\ \midrule
        Uniform & \checkmark && $30.0$ & $40.2$ & $46.8$ & $50.8$ & $54.9$ & $59.5$ \\
\hline
        Random & \checkmark && $23.4$ & $35.4$ & $45.2$ & $49.9$ & $54.5$ & $59.5$\\
        Max-Entropy & \checkmark && $23.4$ & $26.5$ & $34.8$ & $40.7$ & $45.6$ & $51.8$  \\
        Core-set & \checkmark && $23.4$ & $32.1$ & $37.6$ & $41.6$ & $46.5$ & $53.8$ \\
        VAAL & \checkmark && $23.4$ & $35.9$ & $41.8$ & $48.0$ & $54.0$ & $59.4$ \\
        \rowcolor{lightgray!30}{$K$}-means & \checkmark && $\textbf{33.3}$ & $\textbf{42.5}$ & $\textbf{47.9}$ & $\textbf{51.8}$ & $\textbf{55.7}$ & $\textbf{60.2}$ \\ \midrule
        Uniform && \checkmark & $30.8$ & $37.5$ & $40.8$ & $43.1$ & $45.2$ & $48.1$ \\
\hline
        Random && \checkmark & $23.7$ & $33.9$ & $40.2$ & $42.9$ & $45.2$ & $48.0$ \\
        Max-Entropy && \checkmark & $23.7$ & $25.1$ & $29.5$ & $31.6$ & $32.5$ & $33.4$ \\
        Core-set && \checkmark & $23.7$ & $30.3$ & $32.5$ & $33.7$ & $35.3$ & $38.2$  \\
        VAAL && \checkmark & $23.7$ & $34.3$ & $38.2$ & $41.9$ & $44.8$ & $47.8$ \\
        \rowcolor{lightgray!30}{$K$}-means && \checkmark & $\textbf{33.6}$ & $\textbf{41.2}$ & $\textbf{44.2}$ & $\textbf{46.0}$ & $\textbf{47.7}$ & $\textbf{49.8}$ \\ \bottomrule
   \end{tabular}}
    
    \label{tab:moco}
\end{table}

We perform ablation on {$i$}) sampling and {$ii$}) classification backbone initialization and investigate their contributions to $K$-means performance on ImageNet in Table \ref{tab:ablation_study}.

\vspace{.05in}
\noindent{\bf Ablating the sampling backbone.} In Table \ref{tab:ablation_study}, we find that by changing the selection backbone weights to random and keeping the evaluation setting the same, $K$-means performance drops in low budgets.

\vspace{.05in}
\noindent{\bf Ablating the classification backbone.} Table \ref{tab:ablation_study} also presents the key role of SSL pre-trained classification backbone in achieving strong accuracy. When not fine-tuning the randomly initialized backbone, we train a linear layer on the top of frozen random features.

\vspace{.05in}
\noindent{\bf Effect of fine-tuning.} In Table \ref{tab:ablation_study}, we also report the fine-tuning results of all ablation variants. For randomly initialized backbones, we train both feature extractor and linear classifier using SGD optimizer for $100$ epochs with the same learning rate of $0.1$, which is multiplied by $0.1$ in epochs $30$, $60$, and $90$. For SSL pre-trained variants, we apply mean and standard deviation normalization at features before feeding to the linear layer. The optimizer is Adam and a lower learning rate is used for the backbone compared to the linear layer ($10^{-4}$ vs. $10^{-2}$). It is shown in Table \ref{tab:ablation_study} that back-propagating on a pre-trained model with a new objective causes the model to forget previously learned features and a drop in the performance happens.

\subsection{Effect of the Network Architecture and Self-Supervised Pre-training Method}

We change both selection and classification backbone architectures to ResNet-50 and pre-train them on ImageNet with MoCo-v2 \cite{MoCo-v2} for $800$ epochs. We report the experiments in Section \ref{sec:imagenet} with the new setting in Table \ref{tab:moco}. This table shows that the superiority of {$K$}-means sampling to other active learning methods in low budgets is not sensitive to the choice of architecture or SSL pre-training method.

\section{Discussion}
In general, we expect multi-batch active learning algorithms to perform better than single-batch ones since having machine learning and human in the loop reduces the redundancy in the annotated data. However, in our case, single-batch performs better than multi-batch. We hypothesize this happens since our single-batch method finds a large number of clusters that equals the total budget, which causes to represent very small clusters as well. One may improve the multi-batch method by encouraging diversity between iterations of the sampling.

We believe strong performance of $K$-means, especially in low budgets, happens since selected examples by $K$-means represent the categories even better than random examples in Uniform. As the budget size increases, all selection methods converge to the same performance results. Thus, with no annotation information, $K$-means clustering is a strong active learning baseline to achieve an accurate image classifier in very low budgets.

\section{Conclusion}
Most active learning benchmarks assume they have access to a large budget and large labeled seed pool. We believe there is practical need for active learning with smaller budgets. However, the problem is challenging as some categories in image classification may not be presented in the seed. We introduce a very simple baseline for this problem and show that it outperforms state-of-the-art active learning methods in low budgets. Our method leverages the recent progress in self-supervised learning along with simple $K$-means clustering for selecting the images that need to be annotated.
\\ \\
\noindent {\bf Acknowledgment:} 
This material is based upon work partially supported by the United States Air Force under Contract No. FA8750‐19‐C‐0098, funding from SAP SE, and also NSF grant numbers 1845216 and 1920079. Any opinions, findings, and conclusions or recommendations expressed in this material are those of the authors and do not necessarily reflect the views of the United States Air Force, DARPA, or other funding agencies.

% ---- Bibliography ----
%
% BibTeX users should specify bibliography style 'splncs04'.
% References will then be sorted and formatted in the correct style.
%
\small{
\bibliographystyle{splncs04}
\bibliography{main}
}
\clearpage

\appendix
\section{Appendix}
Here, we provide additional details about Sections \ref{sec:imagenet} through \ref{semi-sup}. Table~\ref{tab:appendix_transfer_dset_details} shows the details of fine-grained datasets used in semi-supervised learning and fine-grained evaluation tasks described in Sec \ref{fine_grained} and \ref{semi-sup}. 
% Figure \ref{fig:cat_dist_all} completes Table 6b of the paper by showing the category distribution of $3,000$ ImageNet images selected by both iterative and non-iterative active learning methods. Table \ref{tab:lin_nn_imagenet_lt} compares top-1 linear and nearest neighbor classification results of different sampling methods on an ImageNet-LT pre-trained backbone as a more realistic feature extractor. 

\begin{table}
\vspace{-.3in}
    \caption{\textbf{Fine-grained datasets details.} Training, val, and test split details of the fine-grained datasets used in semi-supervised learning and fine-grained evaluation tasks are listed. For DTD and Flowers, we use the provided val sets. For Aircraft, we sample 20\% of samples per class.}
    \vspace{-.1in}
    \begin{center}
    % \resizebox{\linewidth}{!}{
    \setlength{\tabcolsep}{0.5em}{
    \begin{tabular}{@{}lcccccc@{}}
        \toprule
        \textbf{Dataset} & \textbf{Classes} & \textbf{Train size} & \textbf{Val size} & \textbf{Test size} & Accuracy measure \\
        \midrule
        DTD \cite{cimpoi2014describing} & $47$ & $1,880$ & $1,880$ & $1,880$ & Top-1 \\
        Aircraft \cite{maji2013fine} & $100$ & $5,367$ & $1,300$ & $3,333$ & Mean per-class \\
        Flowers \cite{nilsback2008automated} & $102$ & $1,020$ & $1,020$ & $6,149$ & Mean per-class \\
        \bottomrule
    \end{tabular}
    }
    
    \label{tab:appendix_transfer_dset_details}
    \end{center}
\end{table}

\vspace{-.5in}

\subsection{Category Distribution of Sampled ImageNet}
\begin{figure}
\begin{center}
  \includegraphics[width=0.6\linewidth]{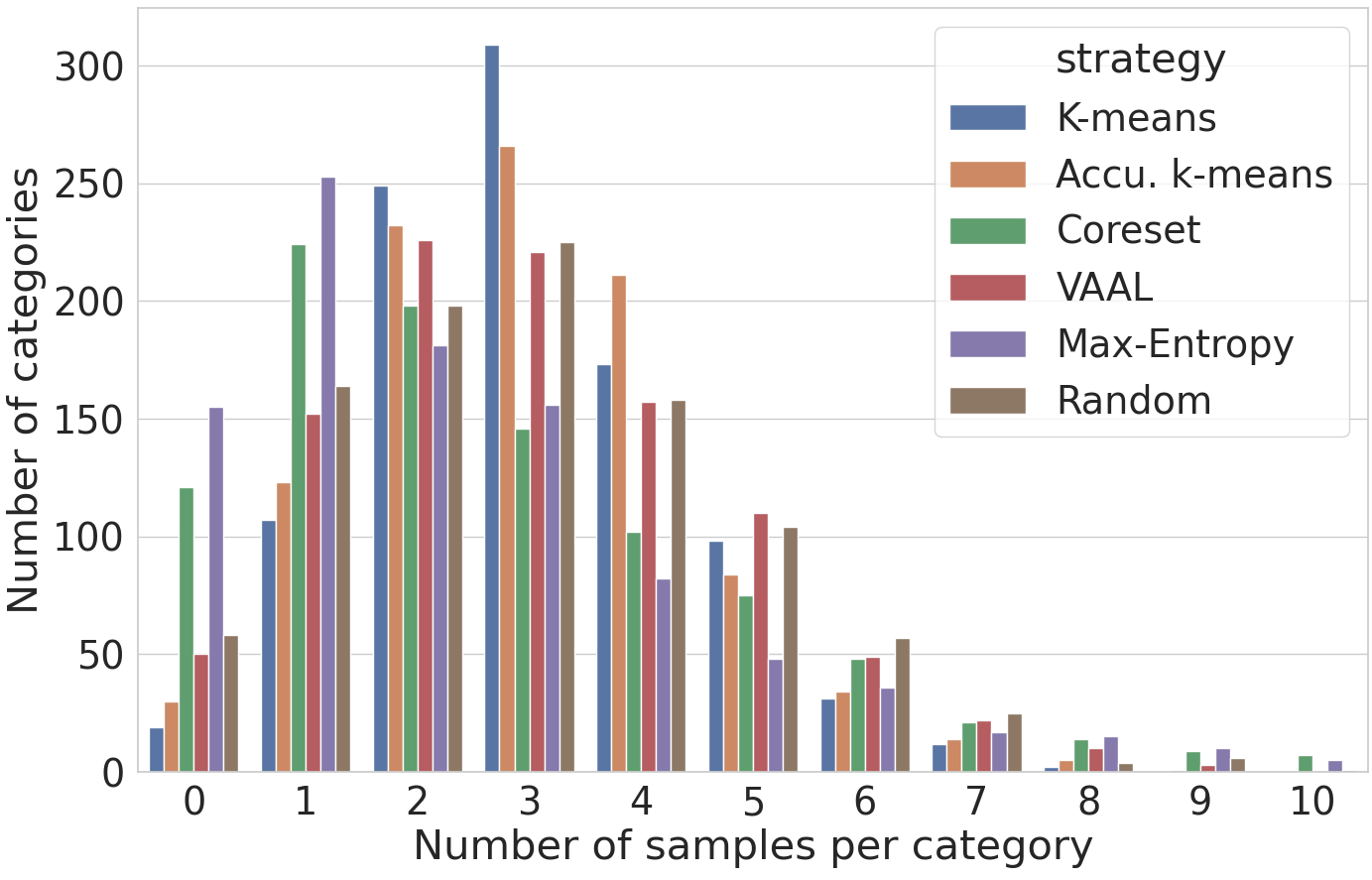}
\end{center}
\vspace{-.2in}
   \caption{{\bf Category distribution of $\textbf{3,000}$ samples of ImageNet ($\textbf{0.2\%}$).}
   Non-peaky category distribution of sampled examples at $3$ for iterative active learning methods that start from random initial pool happens since a random initial pool may not cover all categories and does not provide a strong starting point.}
   \label{fig:cat_dist_all}
 \vspace{-.2in}
\end{figure}

In Figure~\ref{fig:cat_dist_all}, we add the category distribution results of Core-set, VAAL, and Max-Entropy to Table \ref{fig:cat_dist}. This figure illustrates the distribution of ImageNet categories over their number of occurrences in $3,000$ ($0.2\%$) samples of the unlabeled training data. While we prefer the category distribution of selected data points to have a peak at $3$ and be zero otherwise, which will happen with Uniform sampling, it is shown in Figure~\ref{fig:cat_dist_all} that Core-set, VAAL, and Max-Entropy are far from this peaky distribution. We believe this happens since these iterative sampling methods start from a random initial pool that may not be representative of all categories. This incomplete category coverage also propagates to model knowledge and selected batches in next sampling rounds. However, $K$-means sampled examples have sharper category distribution around $3$ which means selected images are distributed more uniformly across the categories.

\subsection{Performance Analysis on ImageNet-LT}
\label{imagenet_lt_appendix}

In Section \ref{imagenet-lt}, we use the same frozen backbone pre-trained on unlabeled ImageNet as the feature extractor and show that $K$-means strategies are strong baselines on ImageNet-LT dataset in low budgets. To assure that $K$-means sampling performance on ImageNet-LT is insensitive to category distribution of the unlabeled training set used for pre-training the backbone, we repeat the experiment in Section 4.3 with a ResNet-50 that is pre-trained on ImageNet-LT using BYOL \cite{byol} and report the results in Table~\ref{tab:lin_nn_imagenet_lt_appendix}. As shown in this table, despite a drop in linear and nearest neighbor classification results of all methods, $K$-means still performs better than non-uniform sampling methods in low budgets in both evaluation metrics without taking advantage of annotations.

We hypothesize that pre-training a self-supervised model on an imbalance training set causes a drop in overall accuracies. Such problem is actively being studied in the community and is out of the scope of this paper.

\begin{table}
\vspace{-.3in}
    \caption{{\bf Top-1 linear (LIN) and nearest neighbor (NN) classification results of different sampling methods on ImageNet-LT.} We use a ResNet-50 that is pre-trained on ImageNet-LT using BYOL \cite{byol} to assure the insensitivity of $K$-means to category distribution of unlabeled training set. With no label information, $K$-means performs better than non-uniform sampling methods in both linear and nearest neighbor classification.}
    \renewcommand{\arraystretch}{0.85}
    % \fontsize{8.0pt}{5.0pt}\selectfont
    \centering
    % \resizebox{\linewidth}{!}{
    \setlength{\tabcolsep}{0.3em}\scalebox{0.81}{
    \begin{tabular}{@{}lcccccccc@{}} \toprule
        && & \multicolumn{6}{c}{\bf Budgets} \\ \cline{4-9}
        \textbf{Method} & \textbf{LIN} & \textbf{NN} & \makecell{\texttt{1K} \\0.8\%} & \makecell{\texttt{3K} \\3\%} & \makecell{\texttt{5K} \\4\%} & \makecell{\texttt{7K} \\6\%} & \makecell{\texttt{9K} \\8\%} &
        \makecell{\texttt{12K} \\10\%} \\ \midrule
        Uniform & \checkmark && $5.34\pm.1$ & $\textbf{10.6}\pm.1$ & $\textbf{13.5}\pm.2$ & $\textbf{15.6}\pm.0$ & $\textbf{17.6}\pm.1$ & $\textbf{19.5}\pm.0$\\
\hline
        Random & \checkmark && $5.04\pm.2$ & $8.60\pm.1$ & $11.4\pm.1$ & $12.8\pm.0$ & $14.1\pm.0$ & $15.6\pm.0$\\
        Max-Entropy & \checkmark && $5.04\pm.2$ & $7.41\pm.0$ & $9.29\pm.1$ & $10.7\pm.0$ & $11.9\pm.1$ & $13.6\pm.1$ \\
        Core-set & \checkmark && $5.04\pm.2$ & $7.77\pm.0$ & $9.66\pm.0$ & $10.8\pm.1$ & $12.0\pm.0$ & $13.7\pm.0$ \\
        VAAL & \checkmark && $5.04\pm.2$ & $8.58\pm.0$ & $10.6\pm.0$ & $12.2\pm.1$ & $13.4\pm.0$ & $14.9\pm.0$ \\
        \rowcolor{lightgray!30} Multi {$k$}-means & \checkmark && $\textbf{6.01}\pm.0$ & $\underline{9.69}\pm.0$ & $11.4\pm.1$ & $12.7\pm.0$ & $13.9\pm.0$ & $15.4\pm.0$ \\ 
        \rowcolor{lightgray!60} {$K$}-means & \checkmark && $\textbf{6.01}\pm.0$ & $9.60\pm.1$ & $\underline{11.7}\pm.0$ & $\underline{13.1}\pm.2$ & $\underline{14.4}\pm.0$ & $\underline{15.9}\pm.1$  \\ 
\midrule
        \rowcolor{lightgray!0} Uniform && \checkmark & $4.81\pm.0$ & $7.03\pm.1$ & $8.21\pm.1$ & $9.02\pm.1$ & $\textbf{9.95}\pm.0$ & $\textbf{10.8}\pm.0$\\
\hline
        Random && \checkmark & $4.42\pm.1$ & $6.40\pm.0$ & $7.64\pm.0$ & $8.22\pm.0$ & $8.85\pm.0$ & $9.55\pm.0$ \\
        Max-Entropy && \checkmark & $4.42\pm.1$ & $4.45\pm.1$ & $4.56\pm.1$ & $4.76\pm.0$ & $5.02\pm.0$ & $5.36\pm.0$ \\
        Core-set && \checkmark & $4.42\pm.1$ & $5.58\pm.0$ & $6.34\pm.0$ & $7.03\pm.0$ & $7.65\pm.0$ & $8.40\pm.0$ \\
        VAAL && \checkmark & $4.42\pm.1$ & $6.33\pm.0$ & $7.26\pm.0$ & $7.95\pm.0$ & $8.51\pm.1$ & $9.12\pm.0$ \\
        \rowcolor{lightgray!30} Multi {$k$}-means && \checkmark & $\textbf{5.48}\pm.0$ & $\underline{7.58}\pm.0$ & $\underline{8.50}\pm.0$ & $\textbf{9.20}\pm.1$ & $\underline{9.74}\pm.1$ & $\underline{10.4}\pm.0$ \\
        \rowcolor{lightgray!60} {$K$}-means && \checkmark & $\textbf{5.48}\pm.0$ & $\textbf{7.61}\pm.0$ & $\textbf{8.64}\pm.1$ & $\underline{9.05}\pm.0$ & $\underline{9.74}\pm.0$ & $10.2\pm.0$  \\ \bottomrule
    \end{tabular}
}
\label{tab:lin_nn_imagenet_lt_appendix}
\vspace{-.3in}
\end{table}

\end{document}